\documentclass[a4paper,twoside]{assets/article}

\usepackage{epsfig}
\usepackage{subcaption}
\usepackage{calc}
\usepackage{amssymb}
\usepackage{amstext}
\usepackage{amsmath}
\usepackage{amsthm}
\usepackage{multicol}
\usepackage{pslatex}
\usepackage{apalike}
\usepackage[bottom]{footmisc}
\usepackage{algorithm2e}
\usepackage{multirow}
\usepackage{tabularx}
\usepackage[printonlyused]{acronym}
\usepackage{hyperref}

\usepackage[english]{babel}
\usepackage{assets/SCITEPRESS}     

\addto\extrasenglish{%
}

\begin{document}

\acrodef{mt}[MT]{Mersenne Twister}
\acrodef{cc}[CC]{Computational Graph}

\acrodef{ml}[ML]{Machine Learning}
\acrodef{dl}[DL]{Deep Learning}
\acrodef{hv}[HV]{Hidden Vector}
\acrodef{rc}[RC]{Reservoir Computing}
\acrodef{vgp}[VGP]{Vanishing Gradient Problem}
\acrodef{pr}[PR]{Neural Network Pruning}
\acrodef{hp}[HP]{Hyperparameter}

\acrodef{bp}[BP]{Backpropagation}
\acrodef{bptt}[BPTT]{Backpropagation Through Time}
\acrodef{gc}[GC]{Gradient Clipping}
\acrodef{wd}[WD]{Weight Decay}
\acrodef{da}[DA]{Data Augmentation}
\acrodef{sgd}[SGD]{Stochastic Gradient Descent}

\acrodef{nn}[NN]{Neural Network}
\acrodef{ann}[ANN]{Artificial Neural Network}
\acrodef{dnn}[DNN]{Deep Neural Network}
\acrodef{ffnn}[FF NN]{Feed Forward Neural Network}
\acrodef{rnn}[RNN]{Reccurent Neural Network}
\acrodef{lstm}[LSTM]{Long Short-Term Memory}
\acrodef{gru}[GRU]{Gated Recurrent Unit}
\acrodef{esn}[ESN]{Echo State Network}
\acrodef{esp}[ESP]{Echo State Property}

\acrodef{gelu}[GELU]{Gaussian Error Linear Unit}
\acrodef{relu}[ReLU]{Rectified Linear Unit}

\acrodef{has}[HaS]{Hide-and-Seek}
\acrodef{vdrre}[VDRRE]{Voronoi decomposition-based random region erasing}

\acrodef{vp}[VP]{VoronoiPatches}
\acrodef{re}[RE]{Random Erasing}
\acrodef{ricap}[RICAP]{random image cropping and pasting}
\acrodef{ssim}[SSIM]{Structural SIMilarity Index}

\title{VoronoiPatches: Evaluating A New Data Augmentation Method}
\author{\authorname{Steffen Illium, Gretchen Griffin, Michael Kölle,\\ Maximilian Zorn, Jonas Nüßlein and Claudia Linnhoff-Popien}
\affiliation{Institute of Informatics, LMU Munich, Oettingenstraße 67, Munich, Germany}
\email{\{steffen.illium, michael.koelle, jonas.nuesslein, linnhoff\}@ifi.lmu.de}
}

\keywords{voronoi patches, information transport, image classification, data augmentation, deep learning}

\abstract{Overfitting is a problem in Convolutional Neural Networks (CNN) that causes poor generalization of models on unseen data. 
To remediate this problem, many new and diverse data augmentation methods (DA) have been proposed to supplement or generate more training data, and thereby increase its quality.
In this work, we propose a new data augmentation algorithm: VoronoiPatches~(VP).
We primarily utilize non-linear re-combination of information within an image, fragmenting and occluding small information patches.
Unlike other DA methods, VP uses small convex polygon-shaped patches in a random layout to transport information around within an image.
Sudden transitions created between patches and the original image can, optionally, be smoothed. 
In our experiments, VP outperformed current DA methods regarding model variance and overfitting tendencies. 
We demonstrate data augmentation utilizing non-linear re-combination of information within images, and non-orthogonal shapes and structures improves CNN model robustness on unseen data.}

\onecolumn \maketitle \normalsize \setcounter{footnote}{0} \vfill

\section{\uppercase{Introduction}}
\label{sec:introduction}
\vspace{-6pt}
Fueled by big data and available powerful hardware, Deep \acp{ann} have achieved remarkable performance in computer vision thanks to the recent development steps. 
But deeper, and wider networks with more and more parameters are data hungry \cite{agga18} beasts.
For a wide variety of problems, a sufficient amount of training data is critical to achieve good performance and avoiding overfitting with modern (often oversized) \acp{ann}. 
Unfortunately, through mistakes in the process of acquiring, mislabeling, underrepresentation, imbalanced classes, etc., data sources can hard to deal with appropriately. \cite{illium2021visual,illium2020surgical}.
In such cases, straight away learning from such real-world datasets might not be as easy as many common research data sets suggest. 
To overcome this challenge, \ac{da} is commonly used alongside other regularization techniques because of its effectiveness and ease of use \cite{shor19}.
Over the years, various methods (e.g., occlusion, re-combination, fragmentation) have been developed and reviewed and contrasted in research.
In the work at hand, we propose a logical combination of such existing \ac{da}-methods: \ac{vp}.

First, we introduce the concept of \ac{da} and Voronoi diagrams in \autoref{sec:preliminaries}, then we introduce and discuss existing and related works in \autoref{sec:related_work}. 
In consequence, we propose our approach (\ac{vp}), the dataset used as well as our experimental setup in \autoref{sec:method}.
Finally, we present the results of our experiments in \autoref{sec:experiments} just before we conclude in \autoref{sec:conclusion}.
\section{\uppercase{Preliminaries}}
\label{sec:preliminaries}
\vspace{-6pt}
\textbf{\acf{da}} is a technique used to reduce overfitting in \acp{ann}, which, in the most extreme cases, is hindered from generalization (the major advantage of \acp{ann}) by perfectly memorizing its training data.
The consequence is poor performance on unseen data.
A function, learned by an overfit model, exhibits high variance in its output \cite{shor19} by overestimating which ultimately leads to poor overall performance.
The amount of a model's variance can be thought of as a function of its size. 
Assuming finite samples, the variance of a model will increase as its number of parameters increases \cite{burn02}.
This is where \ac{da} methods can be applied (on training data) to increase the size and diversity of an otherwise limited (e.g., size, balance) data set.

Such approaches have been very successful in the domain of computer vision (and many others) with the advent of deep CNN \cite{shor19} in the past.
Image data is especially well suited for augmentation, as one major task of \acp{ann} is to be robust to invariance of objects or image features in general. 
\Ac{da} algorithms, on the other hand, are made to generate such invariances.
As many other comparable domains exhibit their own challenges and characteristics, in this work, we restrict ourselves to \ac{da} methods for images.

In general, \ac{da} takes advantage of the assumption that more information can be extracted from the original data to enlarge a data set. 
It follows that a data set supplemented with augmented data represents a more complete set of all possible data, i.e., closing the real-world gap and promoting the ability to generalize.
There are two categories. In a \textit{data warping method}, existing data is transformed to inflate the size of a data set. Whereas, \textit{methods of oversampling} synthesize entirely new data to add to a data set \cite{shor19}.\\

\noindent
\textbf{Voronoi diagrams} are geometrical structures which define the partition of a space using a finite set of distinct and isolated points (\textit{generator points}).
Every other point in the space belongs to the closest generator point.
Thus, the points belonging to each generator point form the \textit{regions} of a Voronoi diagram \cite{okab00}.
We will further focus the explanations and definitions to 2-dimensional Voronoi diagrams (spanning Euclidean space), as used in this work.

Let S be a set of $n \ge 3$ generator points $p, q, r \ldots$ in Euclidean space $\mathbb{R} ^ 2$.
The distance $d$ between an arbitrary point $x = (x_1, x_2)$ and the generator point $p = (p_1, p_2)$ is given as:

\begin{equation}
    d(p, x) = \sqrt{ (p_1 - x_1) ^ 2 + (p_2 - x_2) ^ 2}
\end{equation}

If we examine generator points $p$ and $q$, we can define a line which is mutually equidistant as:

\begin{equation}
    D(p, q) = \{ x | d(p, x) \leq d(q, x) \}
\end{equation}

A Voronoi region belonging to a generator point $p \in S$, $VR(p, S)$, is the intersection of half-planes $D(p, q)$ where $q$ ranges over all $p$ in $S$:

\begin{equation}
    VR(p, S) = \bigcap_{q \in S, q \neq p} D(p, q)
\end{equation}

In other words, Voronoi region of $p$ ($VR(p, S)$) is made up of all points $x \in \mathbb{R} ^ 2$ for which $p$ is the nearest neighboring generator point. 
This results in a convex polygon, which may be bounded or unbounded.
The boundaries of regions are called \textit{edges}, which are constrained by their endpoints (\textit{vertices}).
An edge belongs to two regions; all points on an edge are closest to exactly two generator points. 
Vertices are single points that are closest to three or more generator points.
Thus, the regions of a Voronoi diagram form a polygonal partition of the plane, $V(P)$ \cite{aure91,aure13}.

\begin{figure}[hbt!]
    \centering
    \includegraphics[width=0.5\linewidth]{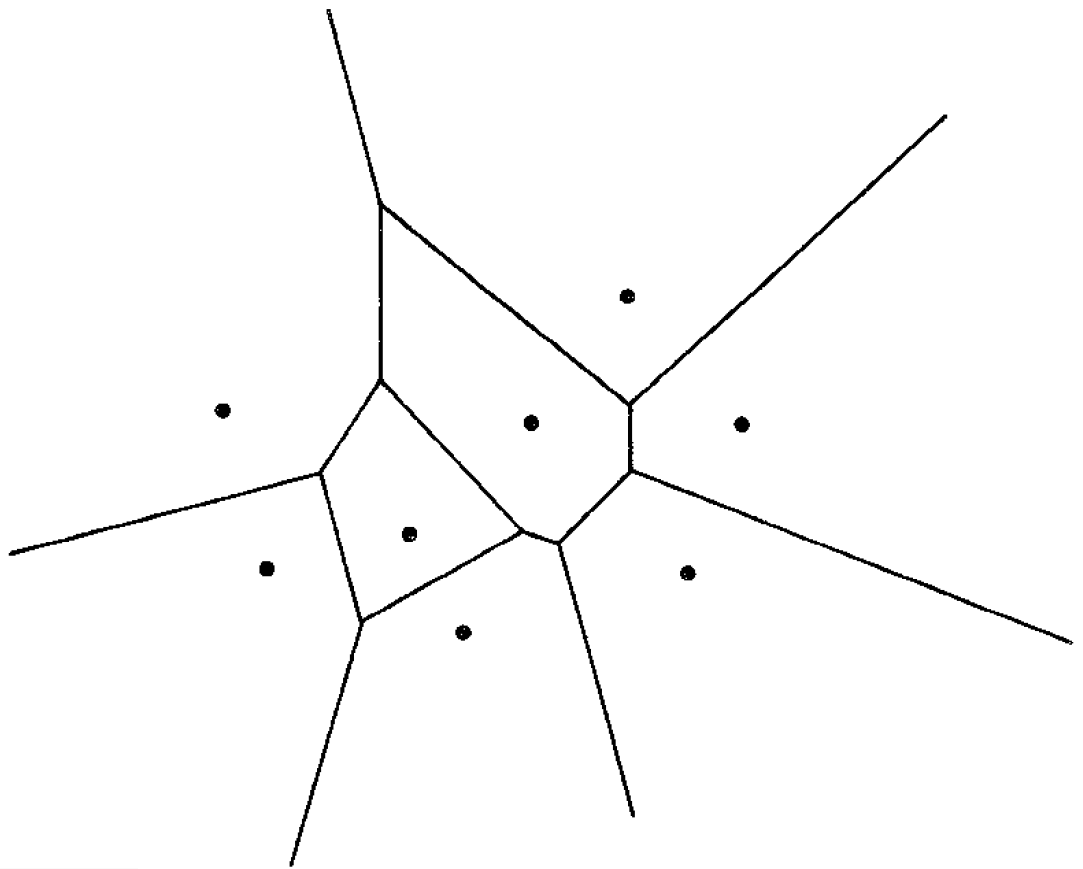}
    \caption{\textit{Voronoi Diagrams:} A simple diagram for eight generator points: $V(P)$ with  $|P| = 8$ \cite{aure91}.}
    \label{fig:vor}
\end{figure}

\section{\uppercase{Related Work}}
\label{sec:related_work}
\vspace{-6pt}
With the preliminaries introduced, we now survey some existing and related work in the field of \acl{da} in the context of deep \aclp{ann}.

\subsection{Occlusion}
\label{sec:occlusion}
\begin{figure}[bt!]
    \centering
    \includegraphics[width=\linewidth]{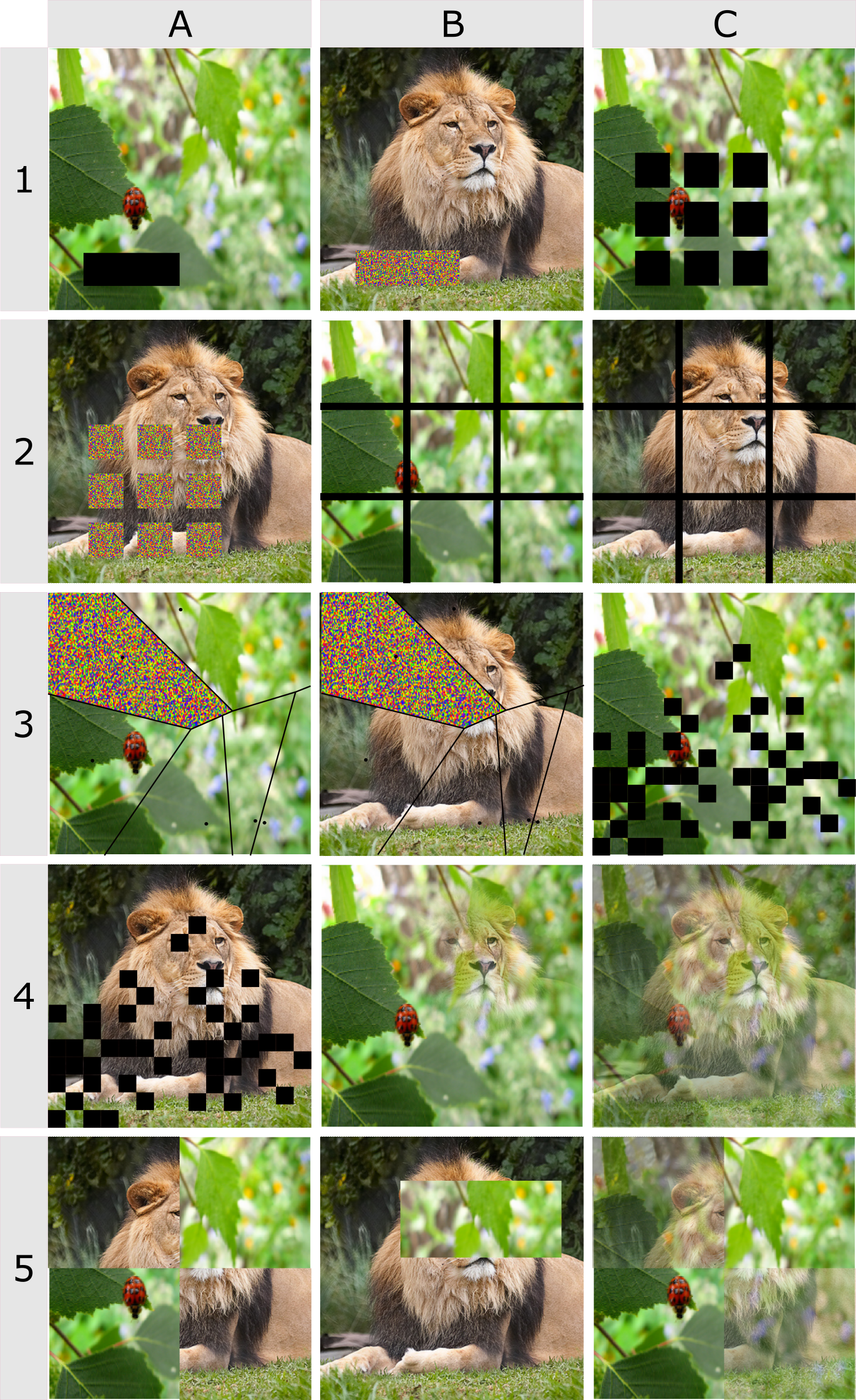}
    \caption{\textit{Combined Augmentations Showcase}: 
                Overview of some available data augmentation methods.
                \textbf{Rows 1-3 + 4A:} Occlusion methods;
                \textbf{4A\&B + Row 5:} (Linear) combination methods.}
    \label{fig:augs_combined}
\end{figure}

\vspace{-6pt}
Methods employing the principle of occlusion mask parts from model input.
In consequence, it encounters more varied combinations of an object's features and its context.
This forces a stronger recognition of an object by its structure. 
Two of the earliest methods that use occlusion are \textit{Cutout} \cite{devr17} and \textit{\ac{re}} \cite{zhon17}. 
Both remove one large, contiguous region from training images (cf.~\autoref{fig:augs_combined}, 1A\&B). 
Through \textit{\ac{has}} \acp{nn} learn to focus on the object overall, occluding parts of the input~\cite{sing18}.
This approach removes more varied combinations of smaller regions in a grid pattern from neighboring model input, however, removed regions may also form a larger contiguous region.
This, on the same hand, is the major limitation of early \ac{da} methods, which can result in the removal of all of an object (or none of it).
\textit{GridMask} \cite{chen20} tries to prevent these two extremes by removing uniformly distributed regions (cf.~\autoref{fig:augs_combined}, 1C, 2A).

While the use of simple orthogonal shapes and patterns is a common characteristic of occlusion-based methods, \ac{vdrre} demonstrates a potential advantage of removing more complex shapes \cite{abay21}.
For the tasks of facial palsy detection and classification, \ac{vdrre} evaluated the use of Voronoi tessellations (cf. Figure\ref{fig:augs_combined}, 3A\&B).

An additional choice for many regional dropout methods is the color of the occlusion mask. Options include random values, the mean pixel value of the data set, or black or white pixels \cite{yun19}. Relevant literature includes \cite{devr17,zhon17,chen20,sing18,abay21}. 
Furthermore, regional dropout methods are not always label-preserving depending on the data set \cite{shor19}. 

\subsection{Re-Combination of Data}
\label{sec:re_combine}
\vspace{-6pt}
\Ac{da} methods which re-combine data, mix training images linearly or non-linearly. 
This has been found to be an efficient use of training pixels (over regional dropout methods) \cite{yun19}, in addition to increasing the variety of data set samples \cite{taka18}. 
However, mixed images do not necessarily make sense to a person \cite{shor19} (e.g., lower section in~\autoref{fig:augs_combined}, 4A\&B + Row 5) and it is not fully understood why mixing images increases performance \cite{summ19}.

Non-linear mixing methods combine parts of images spatially. 
\textit{\Ac{ricap}} \cite{taka18} and \textit{CutMix} \cite{yun19} are non-linear methods that combine parts of two or four training images, respectively. 
Corresponding labels are mixed proportionately to the area of each image used.
\textit{RIACP} combines four images in a two by two grid \cite{taka18}. 
\textit{CutMix} fills a removed region with a patch cut from the same location in another training image (cf.~\autoref{fig:augs_combined}, 5B) \cite{yun19}. 

Linear mixing methods combine two images by averaging their pixel values \cite{shor19}. 
There are several methods that use this approach including: \textit{Mixup} \cite{zhan17}, \textit{Between-class Learning} \cite{toko17}, and \textit{SamplePairing} \cite{inou18} (cf.~\autoref{fig:augs_combined}, 4B\&C,5C). 
Interestingly, \cite{inou18} found mixing images across the entire training set produced better results than mixing images within classes. 
Although images may not make semantic sense, they are surprisingly effective at improving model performance \cite{summ19}. 
A critique on how linear methods introduce “unnatural artifacts”, which may confuse a model, can be found in \cite{yun19}.

\subsection{Fragmentation}
\label{subsec:fragmentation}
\vspace{-6pt}
Fragmentation may occur as a side effect of pixel artifacts introduced by occlusions \cite{lee20} or non-linear mixing of images \cite{taka18}, which create sudden transitions at the edges of removed regions or combined images, respectively.
While occluding and mixing information may prevent the network from focusing on `easy' characteristics, deep \acp{ann} may also latch on to created boundaries of these pixel artifacts. 

In \textit{SmoothMix} \cite{lee20} the transition between two blended images is softened to deter the network from latching on to them. 
Alpha-value masks with smooth transitions are applied to two images, which are then combined and blended. Furthermore, \textit{SmoothMix} prevents the network behavior of focusing on `strong-edges' and achieves improved performance while building on earlier image blending methods like \textit{CutMix}. 

\textit{Cut, Paste and Learn (CPL)} \cite{dwib17} is a \ac{da} method developed for instance detection. 
To create novel, sufficiently realistic training images, objects are pasted on to random backgrounds while the object's edges are either blended or blurred. 
Although the resulting images look imperfect to the human eye, they perform better in comparison to images created manually \cite{dwib17}.

While \textit{SmoothMix} and \textit{CPL} aim to minimize artificially introduced boundaries, \textit{MeshCut} \cite{jian20} uses a grid-shaped mask to fragment information in training images and purposefully introduces boundaries within an image  (cf.~\autoref{fig:augs_combined}, 2B\&C). By doing so, the model learns an object by its many smaller parts; and thereby focuses on broader areas of an object. 
In contrast to \textit{SmoothMix} and \textit{CPL}, \textit{MeshCut} demonstrates that intentionally created boundaries can improve model performance.

\section{\uppercase{Method}}
\label{sec:method}
\vspace{-6pt}
We noticed the benefits from non-trivial edges (horizontal \& vertical) by introducing randomly organized Voronoi diagrams \cite{abay21}, as well as the power of occluding multiple small squares (\ac{has})\cite{sing18}, and opted for a conceptional fusion of these approaches.
From our perspective, there also is the need for a \ac{da} technique which does not occlude wide areas of the image (that may hold the main features), as the occlusion of areas with black ($zero$-values) or Gaussian noise seems not to be fully error-prone.
Therefore, we argue in favor of a new category of \acl{da} methods: \textit{Transport}.
The main idea of transport-based approaches is to preserve features and reposition them in their natural context, in contrast to a replacement by non-informative data.

Combined with the concept of nontrivial edges (Voronoi diagram) and occlusion-of-many \ac{has}, we suggest \acf{vp}. This section now describes our method~(\ac{vp}), the data set used for evaluation, and the experimental setup.

\subsection{\acf{vp}}
\label{subsec:VP_main}
\begin{figure*}[hbt!]
    \centering
    \includegraphics[width=\linewidth]{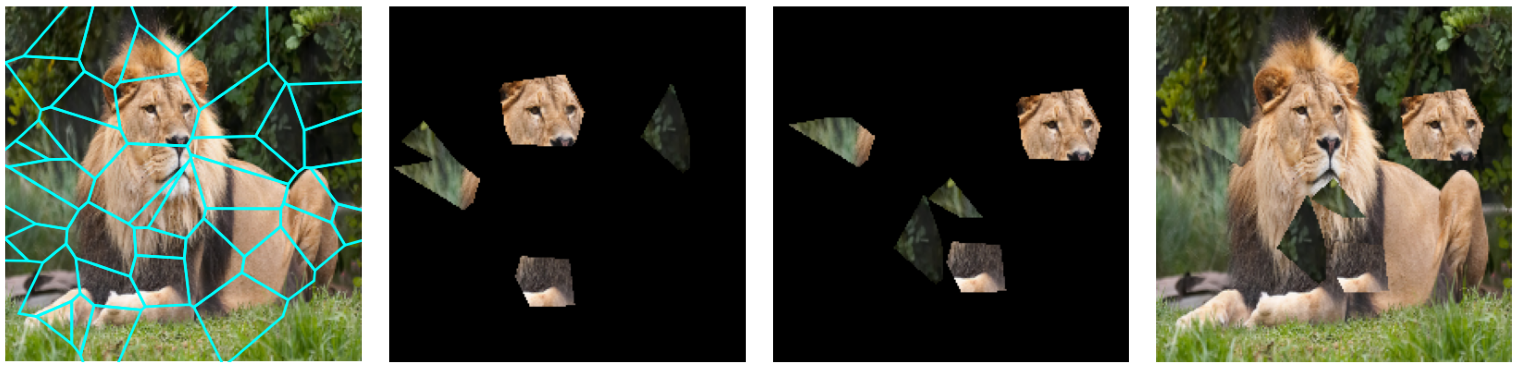}
    \caption{\textit{VoronoiPatches:} An example image and its Voronoi diagram (50 generator points), 5 randomly selected patches \textit{copied}, then \textit{transported} to random locations, and the resulting novel image with sudden transitions (left to right).}
    \label{fig:vor_patches}
\end{figure*}

\vspace{-5pt}
We want to supplement training data with novel images generated online through a non-linear transformation. 
To do so, an image is first partitioned into a set of convex polygons (\textit{patches}) using a Voronoi diagram. 
A fixed number of bounded patches are randomly chosen and copied from the original image. 
These are then pasted randomly over the center of a bounded polygonal region (cf.~\autoref{fig:vor_patches}).
This results in a novel image containing (mostly) the same object features as the original. 
\Ac{vp} may occlude and duplicate parts of the object in the original image while preserving the original label. 
The location, number, and visibility of the parts of an object in an image vary each time \ac{vp} is applied. 
By re-combining information within the image instead of replacing them with random values or a single value, information loss is minimized. 
Due to the random nature of a patch's shape, size, and final location, patches may overlap or not move at all.
This procedure is further described by~\autoref{alg:vor_patch}.

\begin{figure}[hbt!]
    \centering
    \subfloat{\includegraphics[width=.32\linewidth]{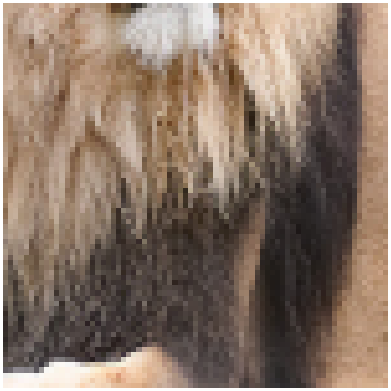}} \hfil
    \subfloat{\includegraphics[width=.32\linewidth]{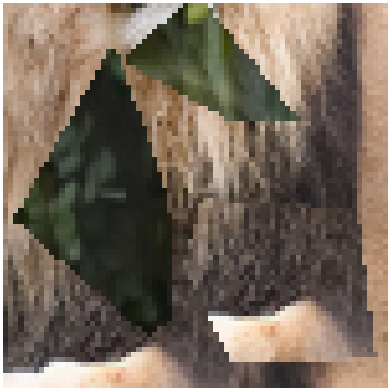}} \hfil
    \subfloat{\includegraphics[width=.32\linewidth]{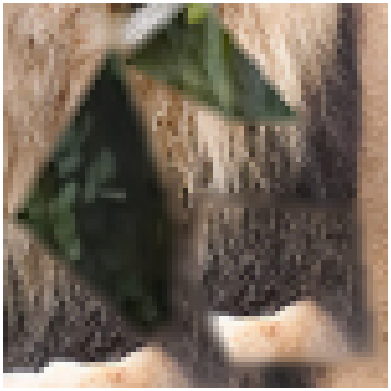}} \\
    \subfloat{\includegraphics[width=.32\linewidth]{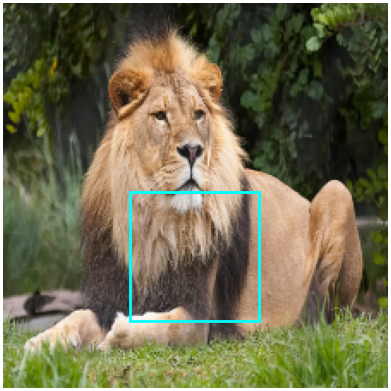}} \hfil
    \subfloat{\includegraphics[width=.32\linewidth]{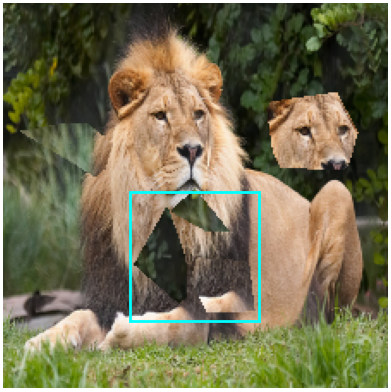}} \hfil
    \subfloat{\includegraphics[width=.32\linewidth]{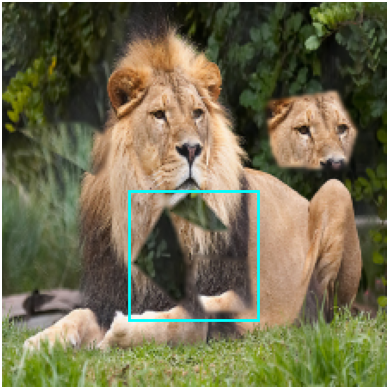}}
    \caption{\textit{Voronoi Patches:} The original image (left), the same image augmented with sudden (middle), or smooth transitions (right).}
    \label{fig:smooth_edges}
\end{figure}

There are three tunable \acp{hp}: \textbf{Number of generator points:} The approximate size and number of patches generated by a Voronoi partition. 
Using few generator points results in fewer larger polygons, and vice versa. 
\textbf{Number of patches}: The number of patches to be transported.
\textbf{Smooth:} The transition style between moved patches and the original image. 
Which may either be left as they are (sudden) or smoothed. 
Sudden transitions may create pixel artifacts caused by sudden changes in pixel values, whereas the application of smooth transitions reduces this effect (cf.~\autoref{fig:smooth_edges}).

\RestyleAlgo{ruled}
\setlength{\algomargin}{0.8em}
\begin{algorithm}[ht]
\SetInd{0.7em}{0.7em}
\caption{VoronoiPatches} 
\label{alg:vor_patch}
\SetKwFunction{KwFn}{VoronoiPatches}
\SetKwProg{Fn}{begin:}{}{end}
\SetKwInOut{Input}{Input}
\SetKwInOut{Output}{Output}
\SetKwComment{Comment}{//}{}
\Input{$ sample $, $ generators $,
       $ patches $, $ smooth $}
\Output{$ aug $}
\Fn{}{
$ aug \gets $ copy of $ sample $

$polygons \gets $ Voronoi(generators)\;
$centroids \gets $ [mean(p) in polygons]\;
    \For{range(patches)}{
        $p \gets$ random($polygons$)\;
        $c \gets$ random($centroids$)\;
        Move $p$ such that arith.Mean($p$) = $c$\;
        \For{$x,y$ in $p$}{
         $aug[x,y] \gets sample[x,y]$\;
            \If{smooth}{
                $s \gets$ gauss.Filter($aug$)\;
                $borders \gets$ \textit{calc\_border()}\;
                \For{$xy_{b}$ in $borders$}{
                    $aug[xy_{b}] \gets$ $s[xy_{b}]$\;
                }
            }
        }
    }
}
\end{algorithm}
\SetInd{0.5em}{1em}

\subsection{Setup \& data set}
\vspace{-5pt}
In this section, we give a brief overview of our data pipeline, and introduce the data set used and performance metrics.
\begin{figure*}[hbt!]
    \centering
    \includegraphics[width=0.85\textwidth]{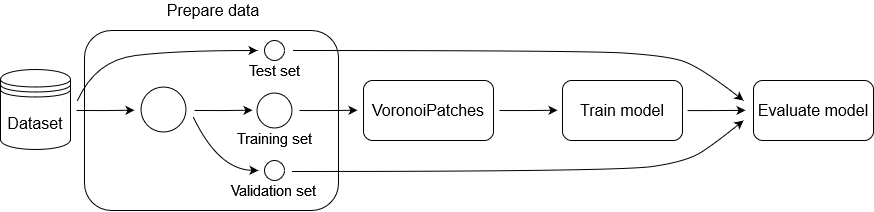}
    \caption{\textit{Data Pipeline}}
    \label{fig:pipeline}
\end{figure*}

\autoref{fig:pipeline} summarizes the procedure used in our data pipeline. 
First, we use an 85:15 split of our training set as an additional validation set for model selection. 
During the training process, performance metrics were calculated and collected each epoch on both the training and validation sets and used to monitor the training process.
Measurements were based on the checkpoint with the highest measured accuracy; please note, that this may over- or underestimate true performance~\cite{agga18}.

We choose the data set based on two factors: sample size and data set size. 
By using medium to large resolution images, we ensure more variation in partitions computed and the ability to generate small enough polygons. 

The 2012 ImageNet Large-Scale Visual Recognition Challenge (ILSVRC) data set\footnote{http://www.image-net.org/}, colloquially referred to as ImageNet, consists of 1.2 million training and 60,000 validation full-resolution images categorized and labeled according to a WordNet\footnote{https://wordnet.princeton.edu/} based class hierarchy into 1,000 classes \cite{russ15}.
While the images' resolution is sufficient, the size of the data set in its entirety is impractical for our experiments.

We used the `mixed\_10' data set\cite{engs19}, which is a subset of the 2012 ImageNet data set (e.g., in~\autoref{fig:diff_size}). 
There are 77,237 training images and 3,000 testing images sampled from the 2012 ImageNet training and validation data sets, respectively. 
These image sets represent ten almost evenly balanced super-classes: dog, bird, insect, monkey, car, cat, truck, fruit, fungus, and boat.

As `mixed\_10' is a well-balanced data set, we choose the classification accuracy in \% as performance metric.
Additionally, we take a look at the variance and entropy introduced by the \ac{da} methods.

\subsection{Experimental Setup}
\label{sec:setup}
\vspace{-5pt}


\noindent\textbf{Baseline:} We first established a baseline model based on SqueezeNet 1.0 (through grid search), which provides us with the default performance metrics.
Based on this model, we apply different \ac{da} methods, in later steps.
We chose SqueezeNet 1.0 because it was designed specifically for multi-class image classification using ImageNet and has a small parameter footprint\cite{iand16}. 
Using 50x fewer parameters, it matched or outperformed the top-1 and top-5 accuracy \cite{iand16} of the 2012 ILSVRC winner, AlexNet, which has approximately 60 million parameters \cite{kriz17}
\begin{figure}[hbt!]
    \centering
    ~\vspace{5pt}
    \subfloat[Insect]{
        \includegraphics[width=0.3\linewidth]{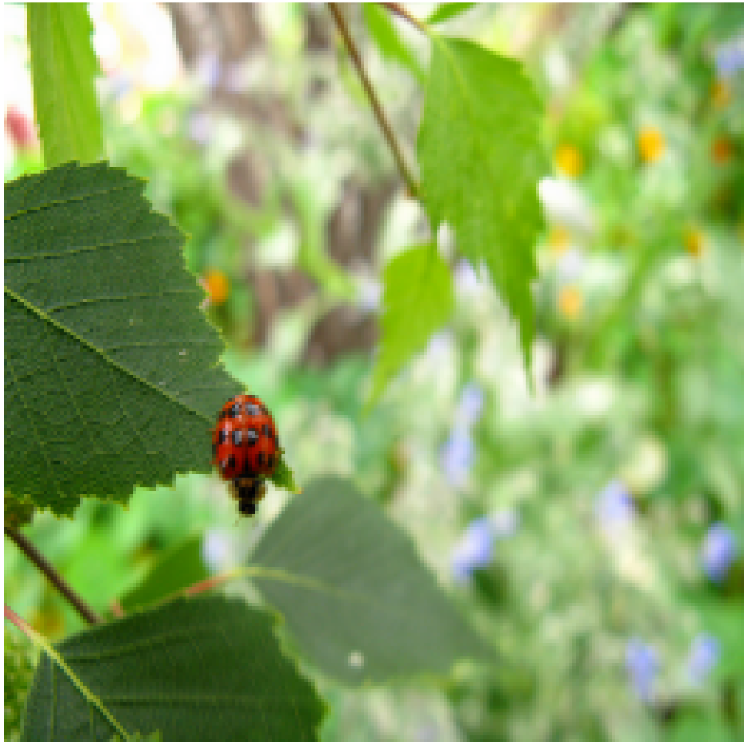}
        \label{fig:obj_sm}}
    \hfil
    \subfloat[Car]{
    \includegraphics[width=0.3\linewidth]{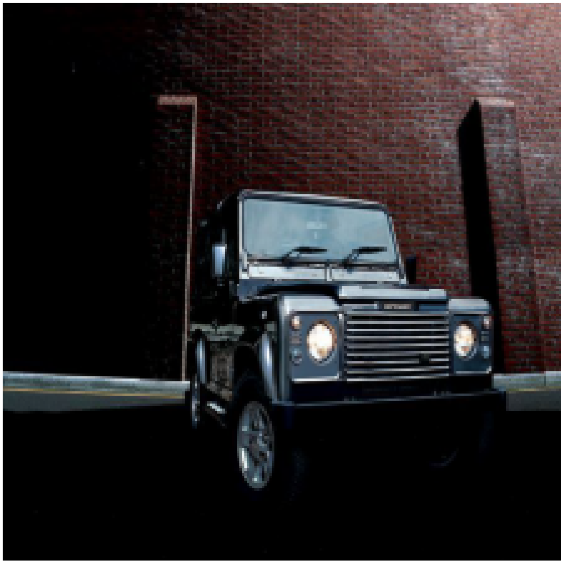}
    \label{fig:obj_med2}}
    \hfil
    \subfloat[Dog]{
    \includegraphics[width=0.3\linewidth]{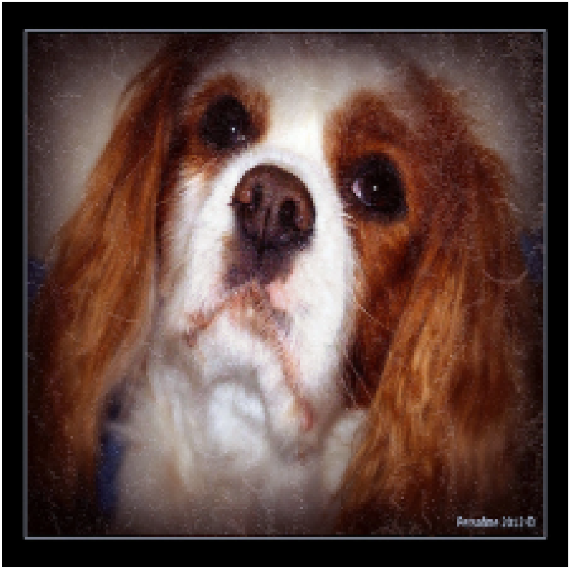}
    \label{fig:obj_lg}}
    \caption{\textit{`mixed\_10' Data Set:} The objects across classes in `mixed\_10' images have a wide range of sizes. Examples from classes (a) – (c) illustrate this.}
    \label{fig:diff_size}
\end{figure}
We only modified the original network architecture by introducing a batch normalization layer as first, to normalize the distribution of our input in a batch-wise fashion.
Since \ac{da} is performed online, this is our only chance to normalize the distribution of our training data set consistently across all models and \ac{da} methods.
Initial model \acp{hp} were adjusted in a `best guess approach' according to literature.
There is no guarantee that we found an optimal configuration; however, our goal was to set up an efficient and well-functioning model for evaluation. 

We further chose cross entropy as a common multi-class classification loss \cite{wang22} and ADAM \cite{king14} as optimizer, which uses adaptive learning rates, which typically require less tuning and converge faster \cite{rude16}. \textbf{Optimizer \acp{hp}:} $lr=0.0001$, $betas=(0.9, 0.999)$, and $eps=1e-08$ with $batch\_size=32$.
As the SqueezeNet 1.0 architecture (using ReLU) requires images at $224 x 224 x 3$ \cite{iand16}, we resized to the fitting resolution and scaled all images channel-wise in the range of $[0, 1]$ using min-max normalization
\footnote{\href{https://scikit-learn.org/stable/modules/generated/sklearn.preprocessing.minmax\_scale.html}{Scikit-learn/Preprocessing: Min-Max-Scale}}. 
By completing this step, the pixel values of our data set have the same scale as the networks' parameters, which may improve convergence time by stabilizing the training procedure \cite{agga18}.
There is no further transformation or augmentation for the baseline model.

\begin{figure}[hbt!]
    \centering
    \hfill
    \subfloat{\includegraphics[width=0.32\linewidth]{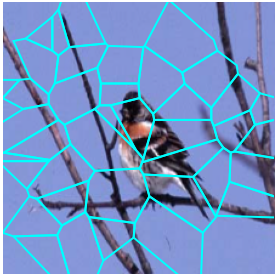}} 
    \hfill
    \subfloat{\includegraphics[width=0.32\linewidth]{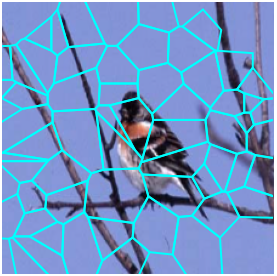}} 
    \hfill
    \subfloat{\includegraphics[width=0.32\linewidth]{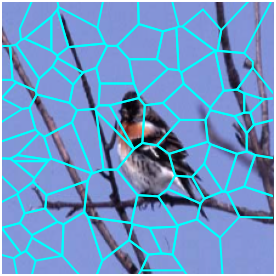}}
    \hfill \\
    \hfill
    \subfloat{\includegraphics[width=0.32\linewidth]{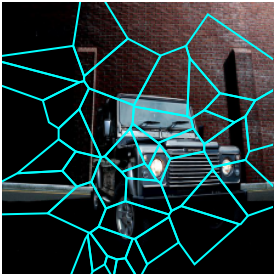}} 
    \hfill
    \subfloat{\includegraphics[width=0.32\linewidth]{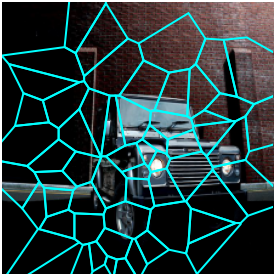}} 
    \hfill
    \subfloat{\includegraphics[width=0.32\linewidth]{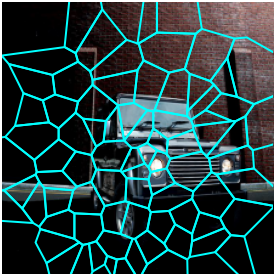}}
    \hfill
    \caption{\textit{Voronoi Diagrams:} Computed with (left) 50, (middle) 70, and (right) 90 generator points.}
    \label{fig:gens}
\end{figure}

Our choice of \acl{vp} \acp{hp} was influenced by the characteristics of our data set, as well as reflects the two main goals: To preserve the label of each augmented image and to occlude or repeat the features of an object or its context in a distributed manner.
The diversity of samples in our data set creates a challenging situation for choosing an optimal number of generator points. 
Specifically, the object-to-context ratio can vary greatly (cf.~\autoref{fig:obj_sm} and \ref{fig:obj_lg}). 
With an image, in which the object is tiny, the danger of all features being occluded by \ac{vp} is still present.
Objects like the car (\autoref{fig:obj_med2}) show an object-to-context ratio, that lies between the extremes of (a) and (c). 
Given the variety of this ratio in the `mixed\_10' data set, it is difficult to estimate what size patch might be too small or large, or how many patches (i.e., total $pixels^2$ moved) are necessary to have a positive impact on model performance. 
To balance the parameter space of our grid search with the variation in object-to-context ratio in our data set, we chose $generators=\{50, 70, 90\}$, cf. \autoref{fig:gens}.
To explore how the size and number of patches might impact performance, we chose $patches=\{5, 10, 15\}$. 
For all combinations, we also considered both transition styles, $smoothing=\{true, false\}$.

\begin{figure*}
    \centering
    \includegraphics[width=\linewidth]{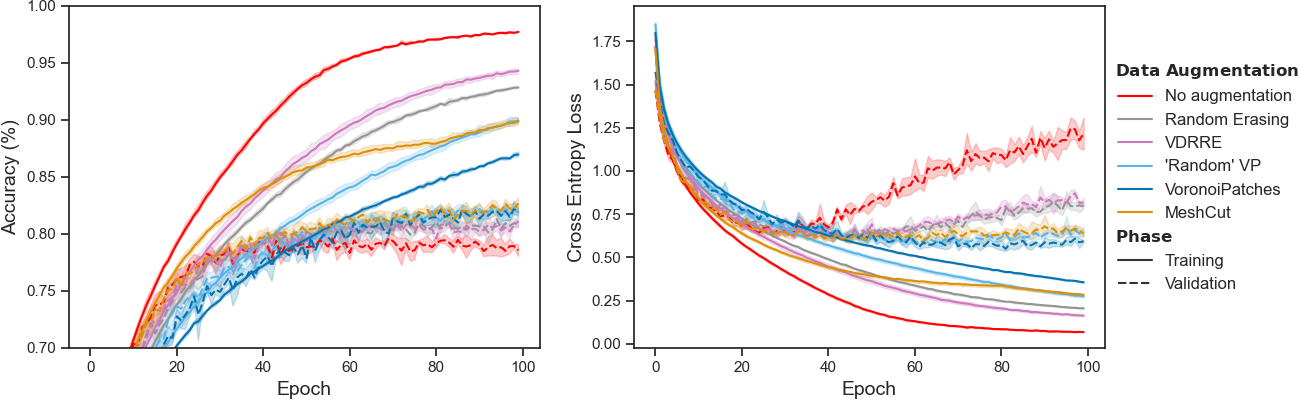}
    \caption{\textit{Avg. Model Performance:}
    \textbf{left:} Train \& val. accuracy over the course of the training (100 epochs). 
    \textbf{right:} CE-loss for both train \& val. data set. Overfitting clearly visible.}
    \label{fig:main_results_incl_overfitting}
\end{figure*}
\section{\uppercase{Experiments}}
\label{sec:experiments}
\vspace{-6pt}
In this section, we describe the course of actions taken to ensure comparable model performance and reproducibility. Finally, we present the observed results of our \ac{vp} model's performance against a same-model baseline (no-DA) and other \ac{da} methods.

\noindent\textbf{Reproducibility:} To ensure the reproducibility of our results, we seeded our interpreter environment, as well as all random number generators involved in the training process.
All models (and \ac{da} methods) along all grid searches use the same seed to ensure reproducibility and validity.
Seeds were chosen at random from $ [0, 2^{32} - 1] $.
To calculate the avg. expected performance, we re-trained our baseline model with the best performing \ac{hp} values from a wide seeded grid search.
After training to convergence (100 epochs, max. acc. 80.1\% at epoch 74), the highest accuracy checkpoints are selected for each seed. 
To find suitable, \ac{vp} \acp{hp} we performed a grid search over all combinations of \textit{generators} and \textit{patches}, w/ and w/o smoothing.
Predefined by the baseline model, the training of \ac{da} methods is limited to 100 epochs.

\begin{figure}[hbt!]
    \centering
    \subfloat{\includegraphics[width=0.32\linewidth]{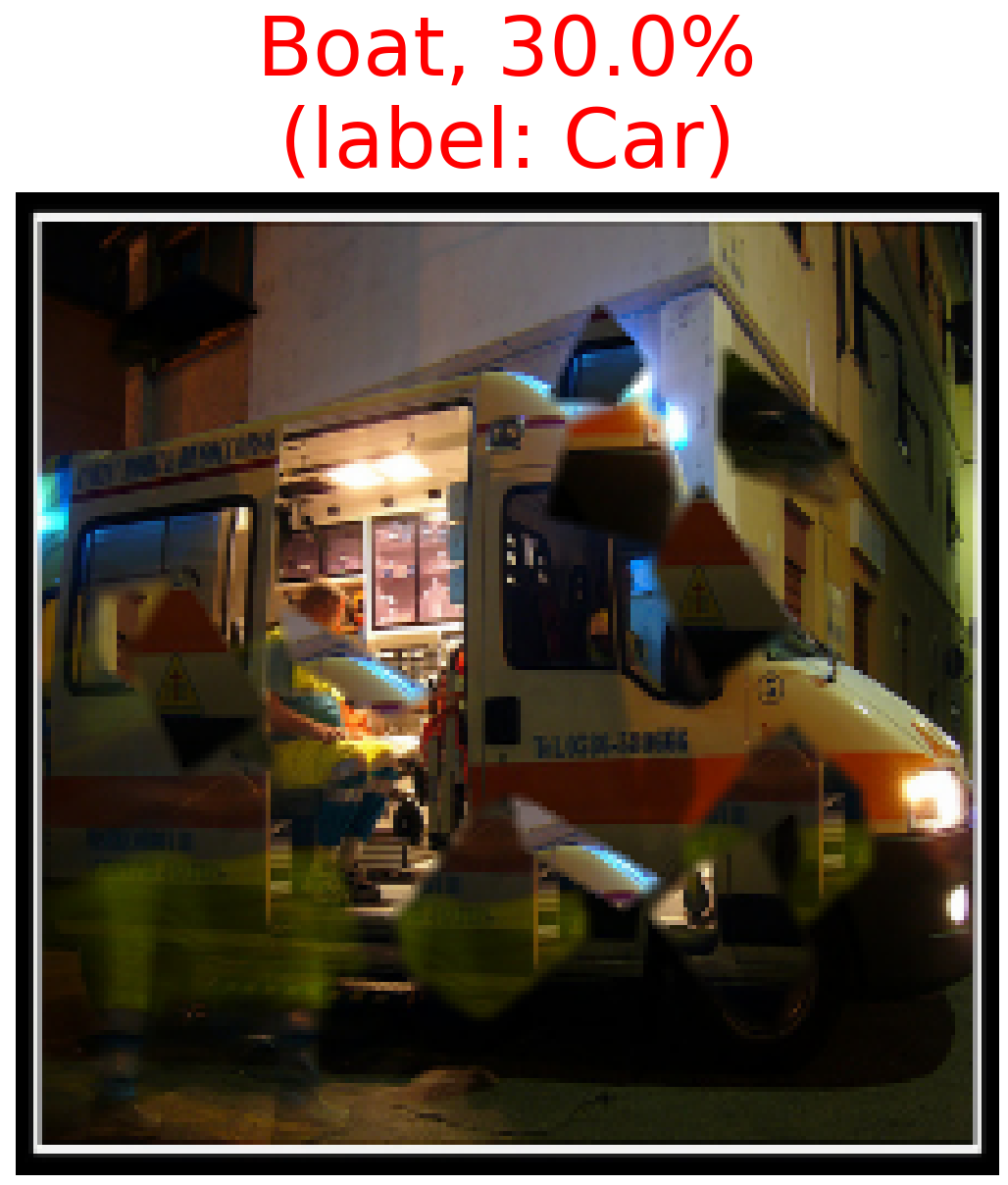}}
    \hfil
    \subfloat{\includegraphics[width=0.32\linewidth]{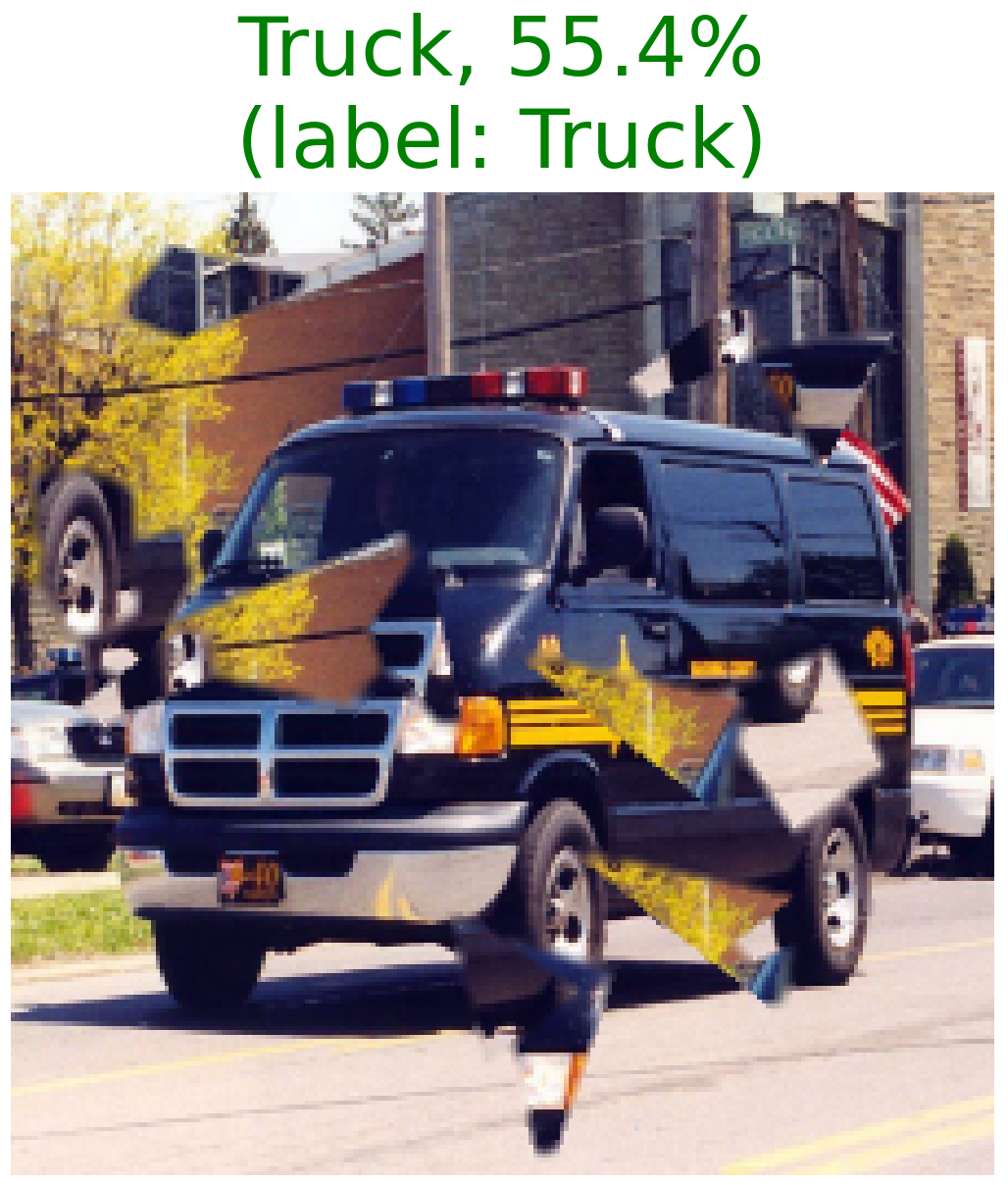}}
    \hfil
    \subfloat{\includegraphics[width=0.32\linewidth]{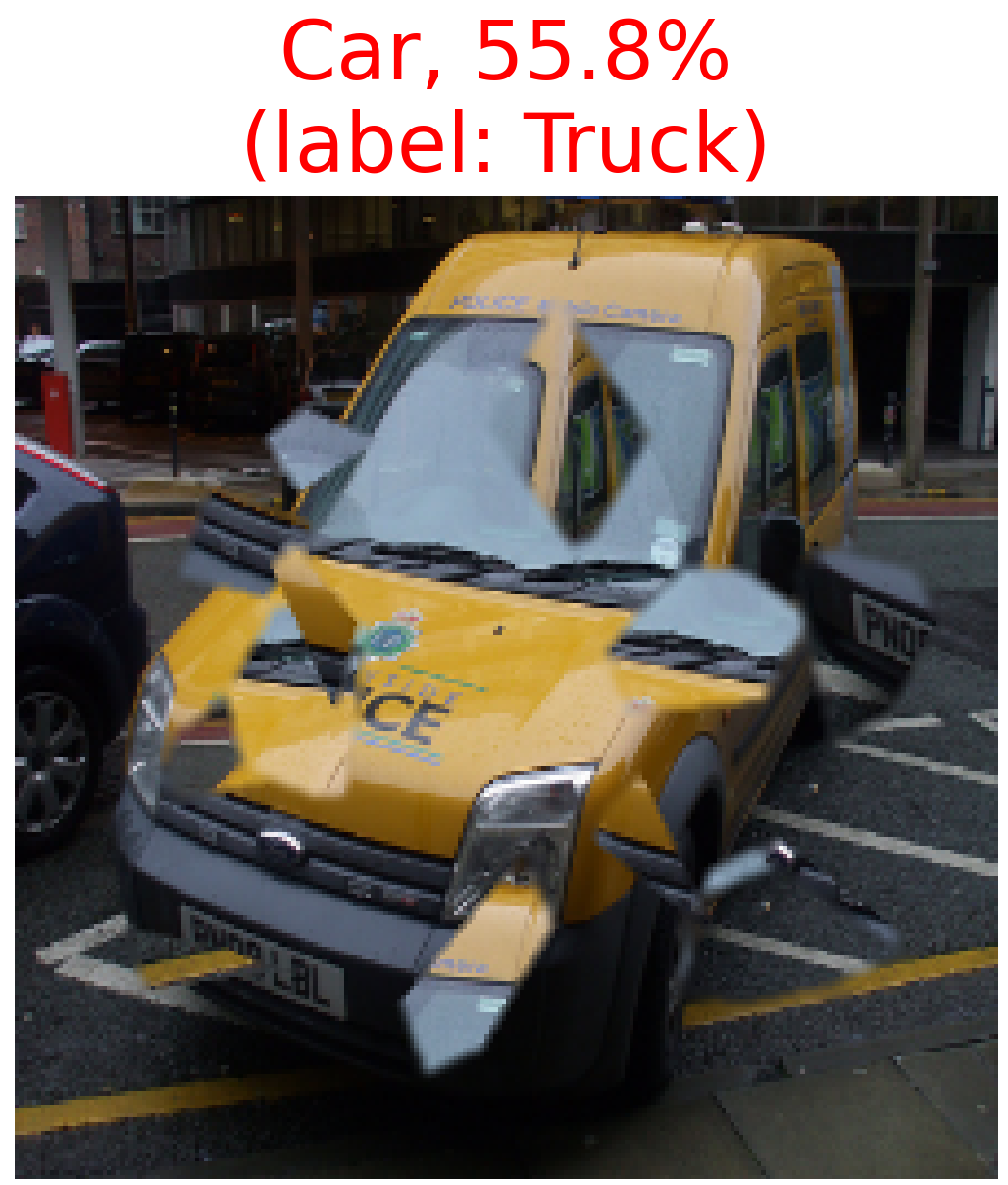}}
    \\
    \subfloat{\includegraphics[width=0.32\linewidth]{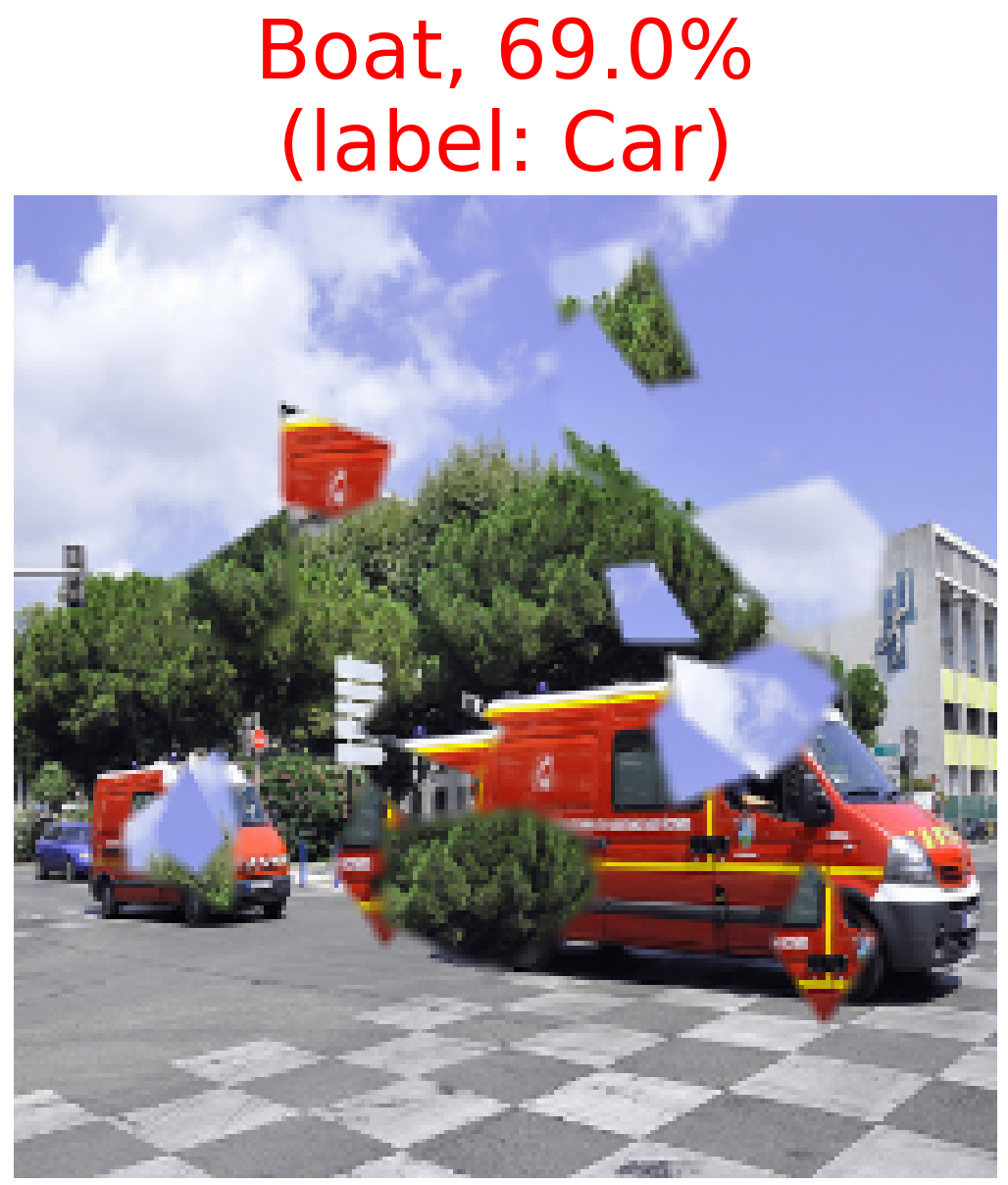}}
    \subfloat{\includegraphics[width=0.32\linewidth]{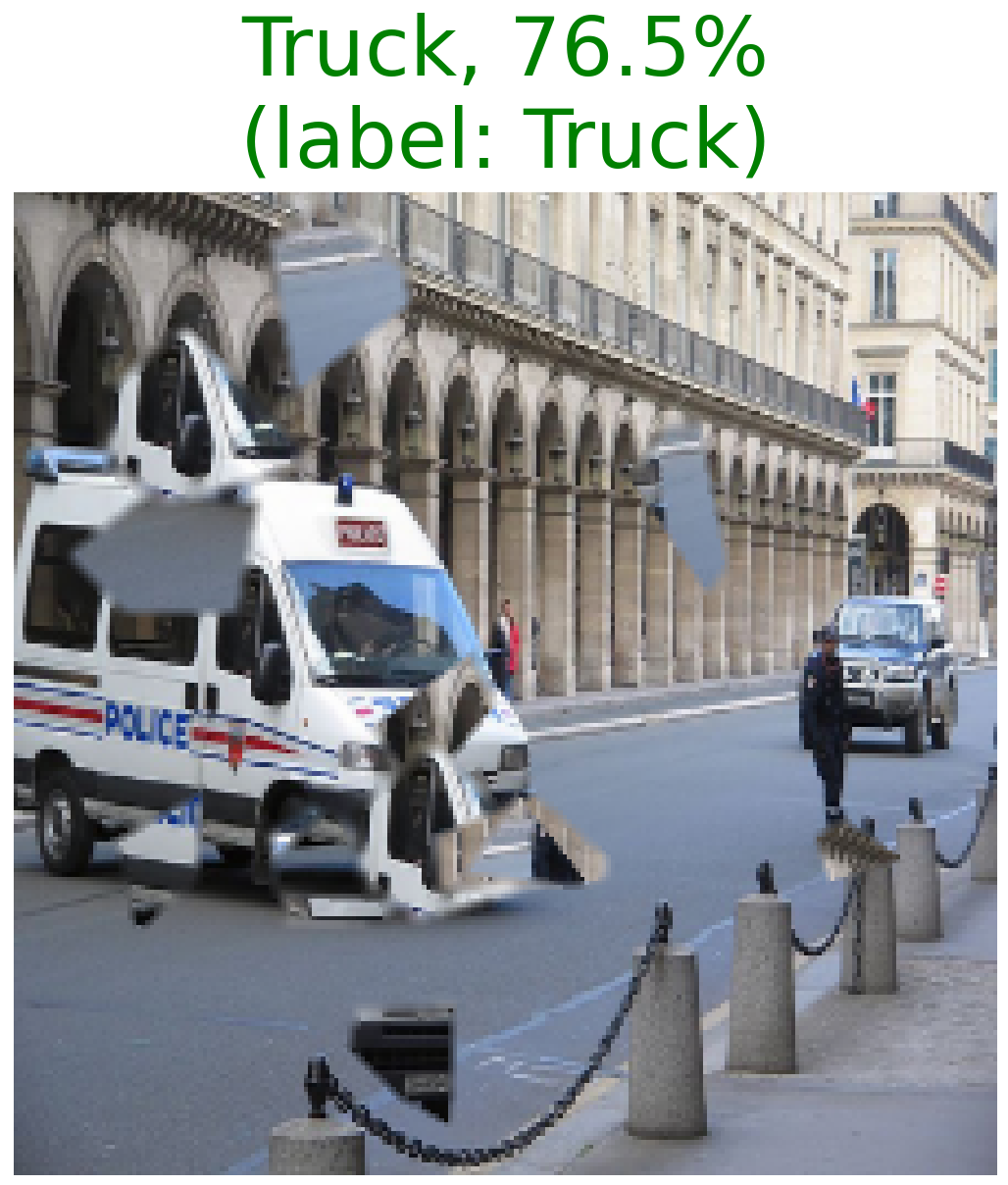}}
    \hfil
    \subfloat{\includegraphics[width=0.32\linewidth]{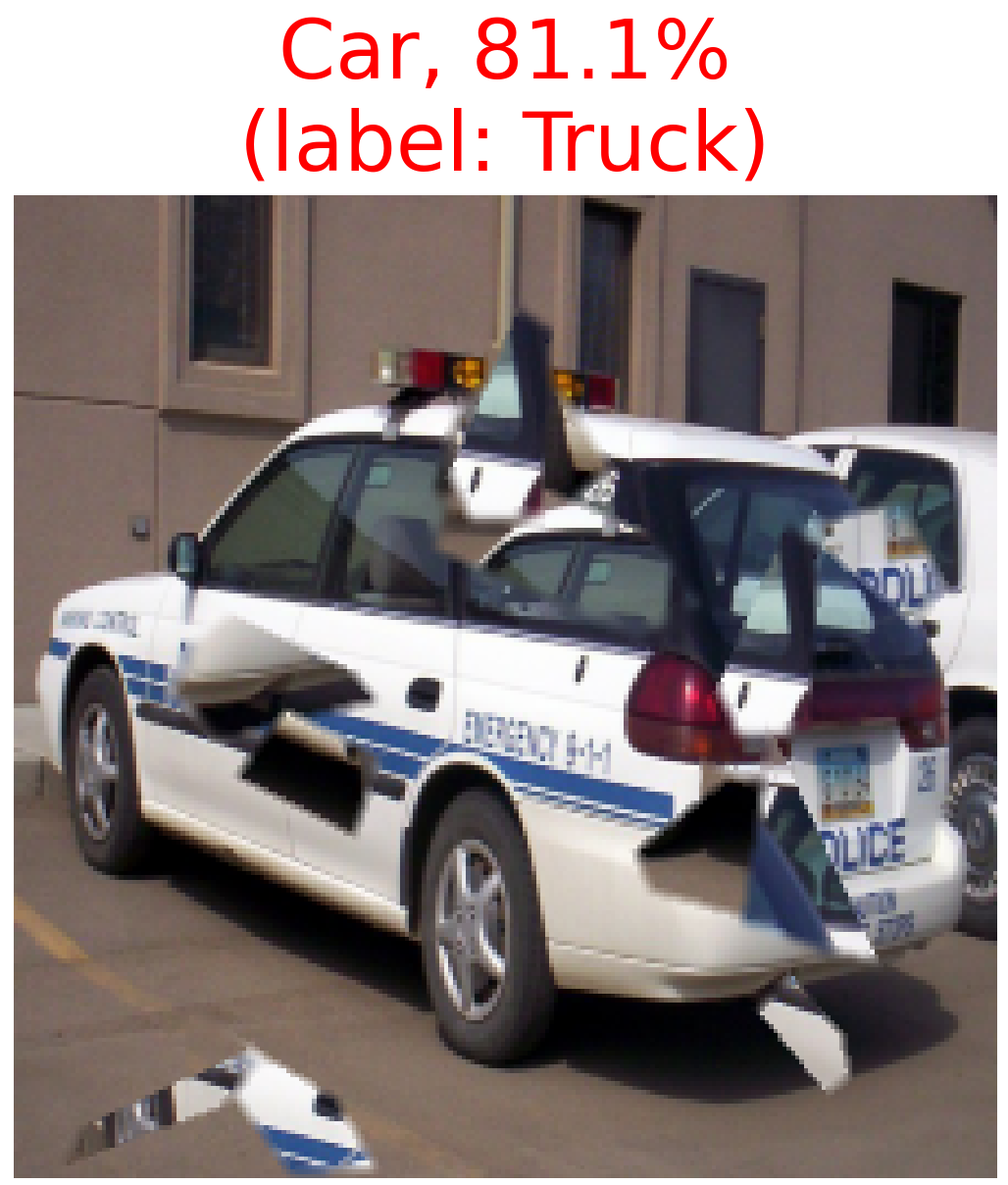}}
    \caption{\textit{Noisy Labels:} Inspection of augmented training images exposed several anomalies. Vans used as ambulances are labeled with `Car' (bottom, left); however, vans used by the police are labeled with `Truck' (middle). A hatchback car used by the police is labeled as `Truck' (bottom, right).}
    \label{fig:noise}
\end{figure}

\noindent\textbf{Results:} For all numbers of patches and patch sizes of \ac{vp} used while training, we observed a performance improvement over baseline performance.
While this improvement is clear, our results showed the amount of improvement varies across \ac{hp} combinations explored.
The highest validation accuracy for \ac{vp} can be reported as 83.6\% at \ac{hp}s: \textit{generators}=70, \textit{patches}=15, and \textit{smooth}=False.
We observed an improvement in val. acc. by about 1.3-3.5\% over the baseline models for all models trained with \ac{vp}. 
On the test set (unseen in training and validation), we evaluated the performance of the resulting best baseline and optimal \ac{vp} models.
The avg. expected performance of our baseline is \textbf{80.9\%} accuracy, 80.9\% macro-recall, and 81\% macro-precision.
The avg. expected performance of a model trained using \ac{vp} is \textbf{83.3\%} accuracy, 83.5\% macro-precision, and 83.3\% macro-recall.
Even though macro-measures are quite similar, the performance of individual classes varies as expected.
However, the classes `Car' and `Truck' stand out as the most difficult for all models to separate, as both perform approximately 10-30\% worse than the others. 
\autoref{fig:noise} shows identified occurrences of noisy `Car' and `Truck' labels among these collected images (\ac{vp} applied).
We cannot rule out that the large performance difference of these classes, on avg., is due to these noisy labels.
During our research, we learned, that other researchers had problems with this situation, which is why ImageNet is currently in debate \cite{beyer2020we}.

\noindent\textbf{Smoothing:}
Unfortunately, we did not observe any clear improvement in the performance of \ac{vp} using smooth over sudden transitions (smooth=false). 
Although the best performing models use sudden transitions, based on our observations in model training; however, we cannot rule out a possible advantage at this time.
Depending on the other \acp{hp}, either approach can be of advantage.
We implemented the width of the transition between patch and image with a fixed factor. In future works, this could be introduced as another \ac{hp}. Both smoothing and the factor could also be bound to a stochastic process (sample on every \ac{vp} application).

\noindent\textbf{Comparisons:}
To verify the validity and utility of \ac{vp}, we evaluate the performance of the \ac{re}, \ac{vdrre} and a combination of both, using the same procedure as outlined above. 
For \ac{re} \acp{hp} outlined in \cite{zhon17}, pixel values `black' (0) and `random' (uniform [0,1)) are chosen. 
\autoref{fig:main_results_incl_overfitting} shows an avg. acc. improvement by \ac{vp} over \ac{re} of 0.6\%-0.7\%.
\acp{vdrre} parameters are chosen as suggested \cite{abay21}.
They established six regions, of which one was then occluded with noise.
On our data set, \ac{vdrre} achieved results, which are close to \ac{re}.
\textit{MeshCut} (augmentation schedule in training and \acp{hp} as stated by the authors \cite{jian20}) behaved worse than `Random'\ac{vp}.
Inspired by the occlusion with Gaussian noise, we also tested a \ac{vp} version with randomly filled patches, rather than transporting sections of the image (more or less comparable with \ac{vdrre}, but at smaller scale).
`Random'\ac{vp} achieved better results than both, \ac{vdrre} and \ac{re}, while performing close to \ac{vp}.

\begin{figure}[hbt!]
    \centering
    \includegraphics[width=\linewidth]{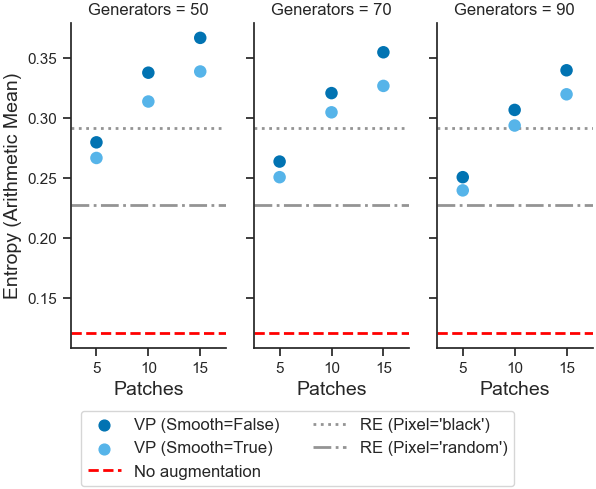}
    \caption{\textit{Entropy}: Avg. effect of Voronoi-Patches (dark blue=sudden, light blue=smooth edges) and \ac{re} (gray lines) measured in entropy $H$ (cf.~\autoref{eq:entropy_h}).}
    \label{fig:entropy_scores}
\end{figure}

\noindent\textbf{Entropy $H$:}
To measure the effect of our method on the training data, we employed entropy $H$ \cite{good16} involving the resulting prob. distribution of each image $x$ (with $P:=prob.$, $K:=classes$):
\begin{equation}
    H(x) = - \sum_{k \epsilon K} P(k) * log(P(k))
    \label{eq:entropy_h}
\end{equation}
We then aggregated, the entropy scores for all images with the arithmetic mean (cf.~\autoref{fig:entropy_scores}).
The avg. entropy is a measure of how balanced the probability distributions are on avg. \cite{good16}. 
For a probability distribution of 10 values (for 10 classes) the lower and upper bounds for entropy are 0.00 and 3.32, respectively. 
The bounds correspond to the case of a perfect classification with 1.0 probability for one class and 0.0 for all others (low entropy) and the case of perfectly balanced probability of 0.1 across all classes (high entropy).
As anticipated, the more data transported within an image by \ac{vp} the more uncertainty is embedded in the model's output (i.e., the more balanced the model's probability distributions are). 
We observe that this effect is greater among all \ac{vp} \ac{hp} combinations than \ac{re} with either `black' or `random' pixels.
We attribute the reduced entropy of \ac{re} using `random' pixels over `black' pixels to the model finding patterns in these values.

\begin{figure}[hbt!]
    \centering
    \includegraphics[width=\linewidth]{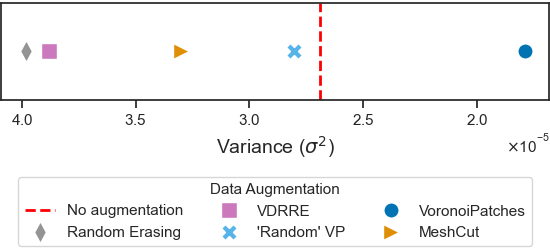}
    \caption{\textit{Model Variance Compare:} Avg. test variance (std) of \ac{vp} and the other \ac{da} methods.}
    \label{fig:vp_re_var}
\end{figure}

\noindent\textbf{Main Findings:} 
\noindent\textit{(1)}~\autoref{fig:vp_re_var} shows measured variance for considered \ac{da} methods.
Over all seeded runs, our proposed method (\ac{vp}) exhibits the lowest variance, even lower than training on non-augmented training data.
By considering the results of the former entropy analysis, especially the last point is quite a surprising find.

\noindent\textit{(2)} By observing the avg. acc. and the CE-loss in~\autoref{fig:main_results_incl_overfitting} we observe a second interesting aspect.
All training runs (w/ and w/o \ac{da}) came at the cost of still overfitting to the training data of `mixed\_10'.
Our method \ac{vp}, among all training runs, showed the least amount of overfitting.
This can be observed by the growing distance of the CE-loss in \autoref{fig:main_results_incl_overfitting} (right).

\noindent\textbf{Further Analysis:} The number of generator points determines the avg. patch size. 
Evaluating the impact of this \ac{hp}, we collected the avg. $pixels^2$ (2D area) for bounded polygons using 50, 70, and 90 generator points. 
The avg. patch sizes are 932, 673, and 528 $pixels^2$, respectively.
We observe that using many smaller patches (15 patches, 528 $pixels^2$), as well as using fewer larger patches results in better model performance (5 patches, 932 $pixels^2$), on avg. 
As we observe no clear improvement using smooth transitions, we assume, either, that the transition areas' width is not optimal, or the pixel artifacts have a positive effect similar to the mesh mask used in \cite{jian20}.

\begin{figure} [hbt!]
    \centering
    \vspace{-4pt}
    \includegraphics[width=\linewidth]{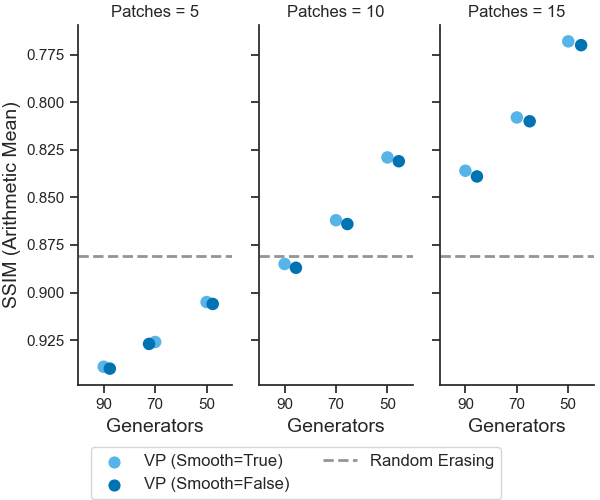}
    \caption{Avg. SSIM for \ac{vp} and \ac{re}.}
    \label{fig:all_ssim}
\end{figure}

To measure how different \ac{hp} values change our original images, we calculate the \ac{ssim} between orig. and augmented images. 
SSIM was developed for the measurement ([0-1]) of image degradation by comparing the luminance, contrast, and structures between two images \cite{wang05}.
$SSIM(original, augmented)=1.0$ represents the trivial case of a perfect match (`identical' images) \cite{wang05}.
Collecting and aggregating (arithmetic mean) \ac{ssim} values for 1,000 iterations with several \ac{hp} combinations \autoref{fig:all_ssim} confirms that by increasing the number of patches or decreasing the number of generators, the similarity of original and augmented images decreases.
The relationship between the avg. \ac{ssim} and avg. entropy can be reported as approx. linear. 
As the structural difference in augmented images increases, so does the entropy of the corresponding model output for these images. 

Further, we observe a drop-off in performance when more than approx. 10,000 total $pixels^2$ are transported. 
The best performance is achieved by 10,095 total $pixels^2$ ($generators=70, patches=15, smooth=False$), a re-combination of 20.1\% of an images' data. 
%
%
There appears to be an optimal avg. patch size and/or number of patches moved similar to the optimal ratio between the grid mask of uniformly distributed squares \cite{chen20}, which controls how many squares make up the mask, as well as how much space is left between.
As a point of reference, the avg. rectangle size generated by \acf{re} is 10,176 $pixels^2$.
In accordance with the default values \cite{zhon17}, \ac{re} was applied with a probability of 50\%. 
On avg., \ac{re} removes a rectangle equal to 20.3\% of an image when applied, whereas \ac{vp} re-orders patches equal to 20.1\%, combined, in each image.
We noticed an improvement in performance by re-combining many small \ac{vp} patches over randomly erasing one large contiguous rectangle with 50\% probability (\ac{re}).

\begin{figure}[hbt!]
    \centering
    \vspace{-4pt}
    \subfloat{\includegraphics[width=0.32\linewidth]{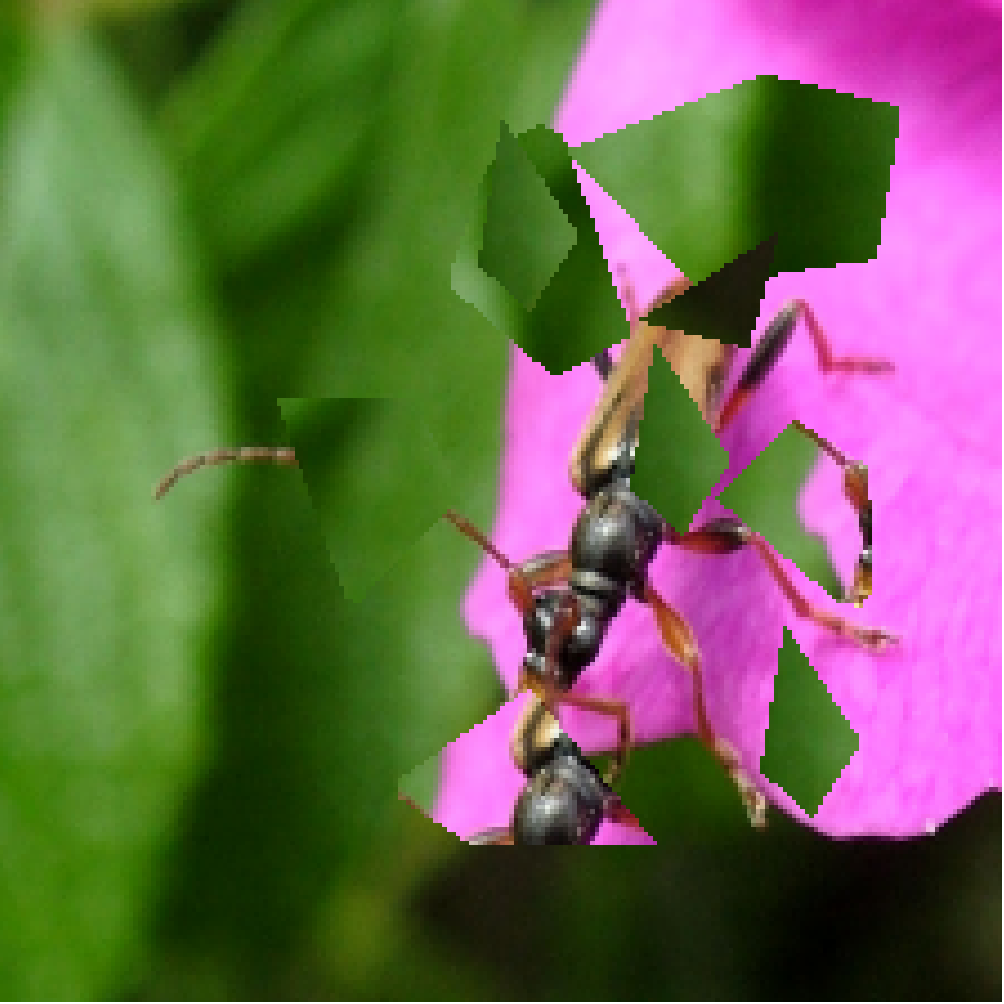}\label{fig:patch_ex_a}}
    \hfil
    \subfloat{\includegraphics[width=0.32\linewidth]{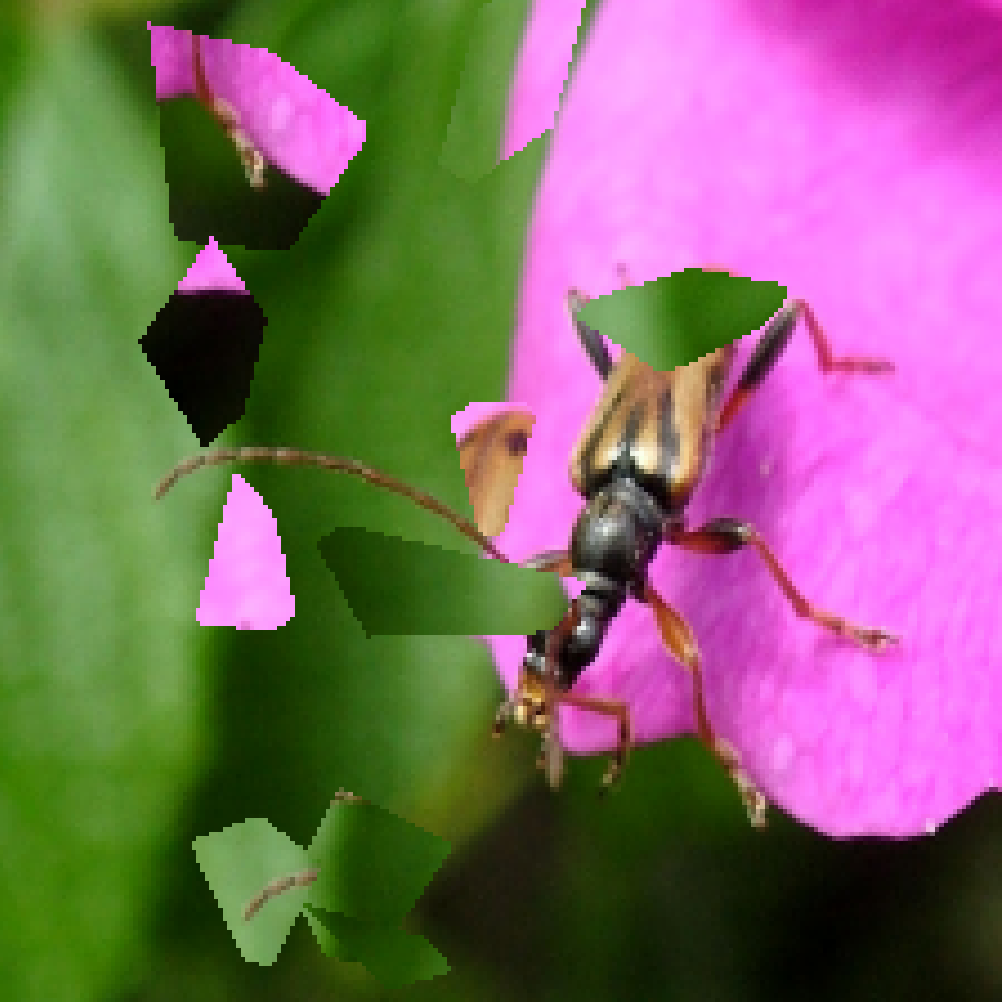}\label{fig:patch_ex_b}} 
    \hfil
    \subfloat{\includegraphics[width=0.32\linewidth]{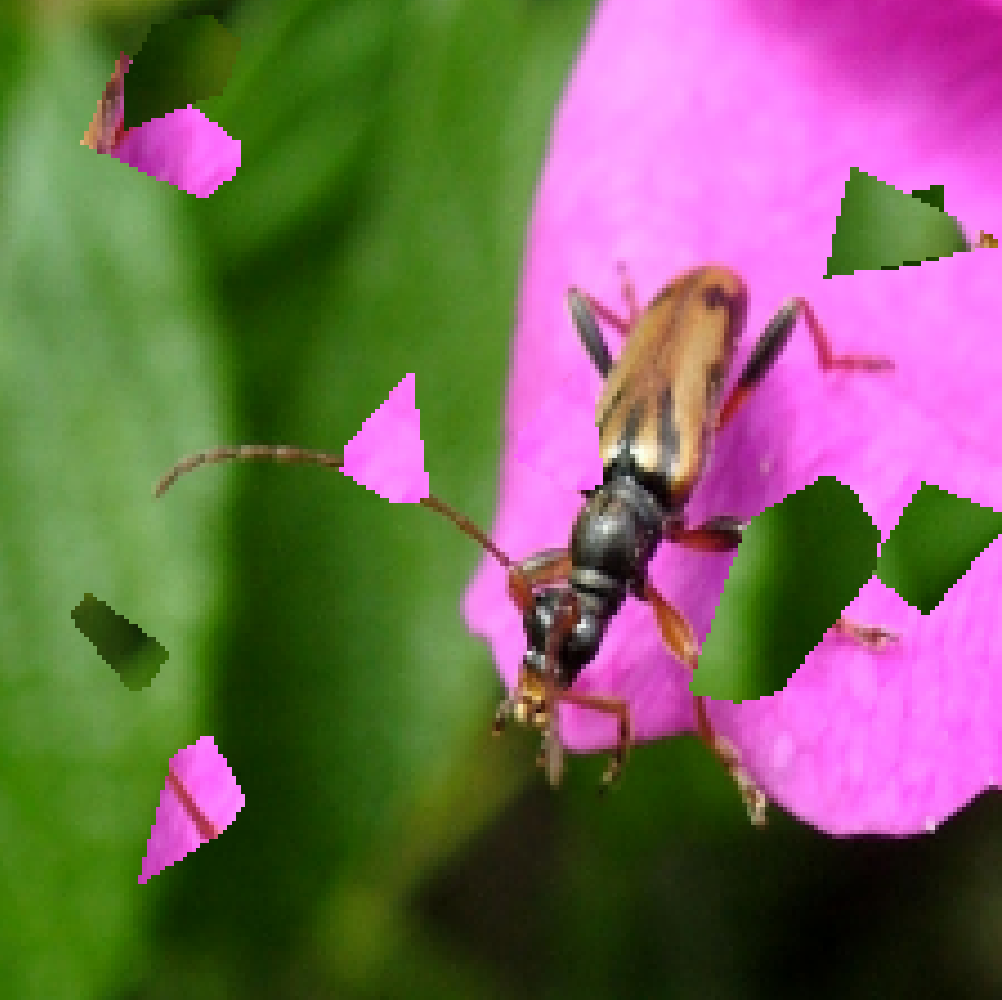}\label{fig:patch_ex_c}} 
    \\
    \subfloat{\includegraphics[width=0.32\linewidth]{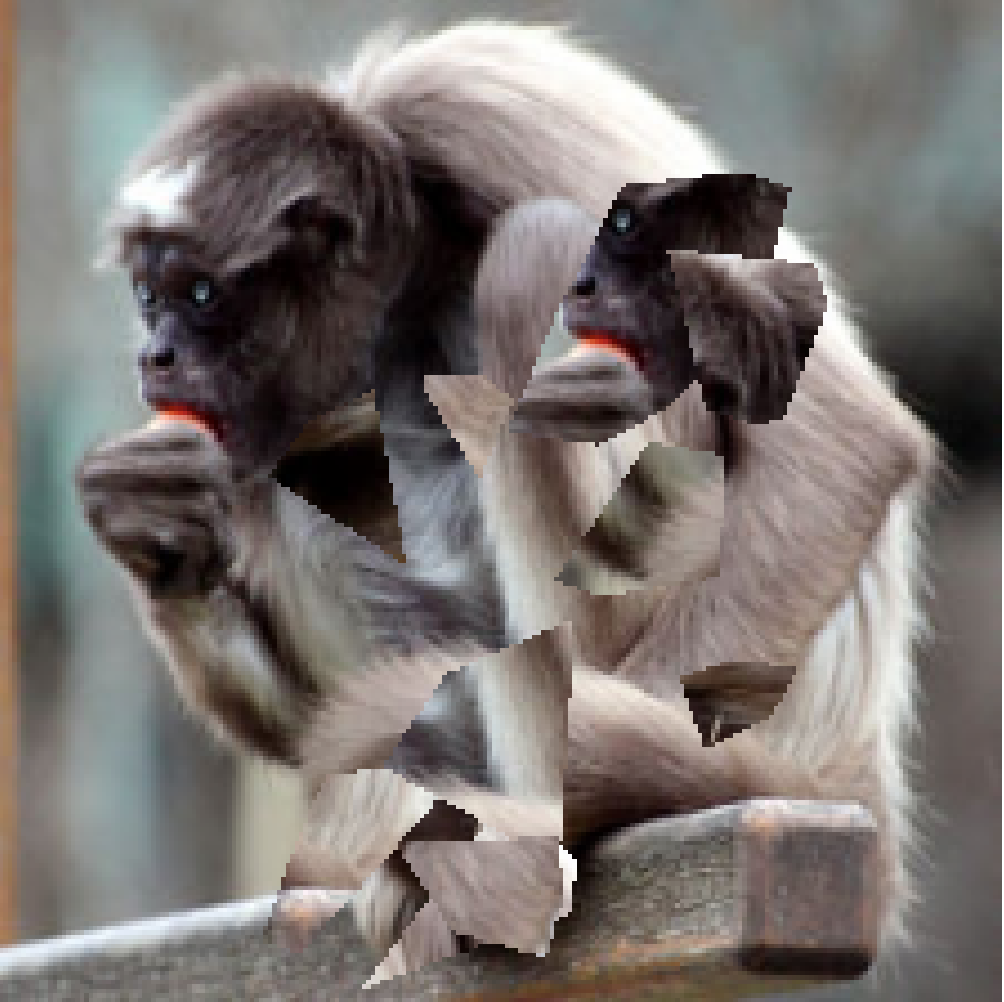}} \hfil
    \subfloat{\includegraphics[width=0.32\linewidth]{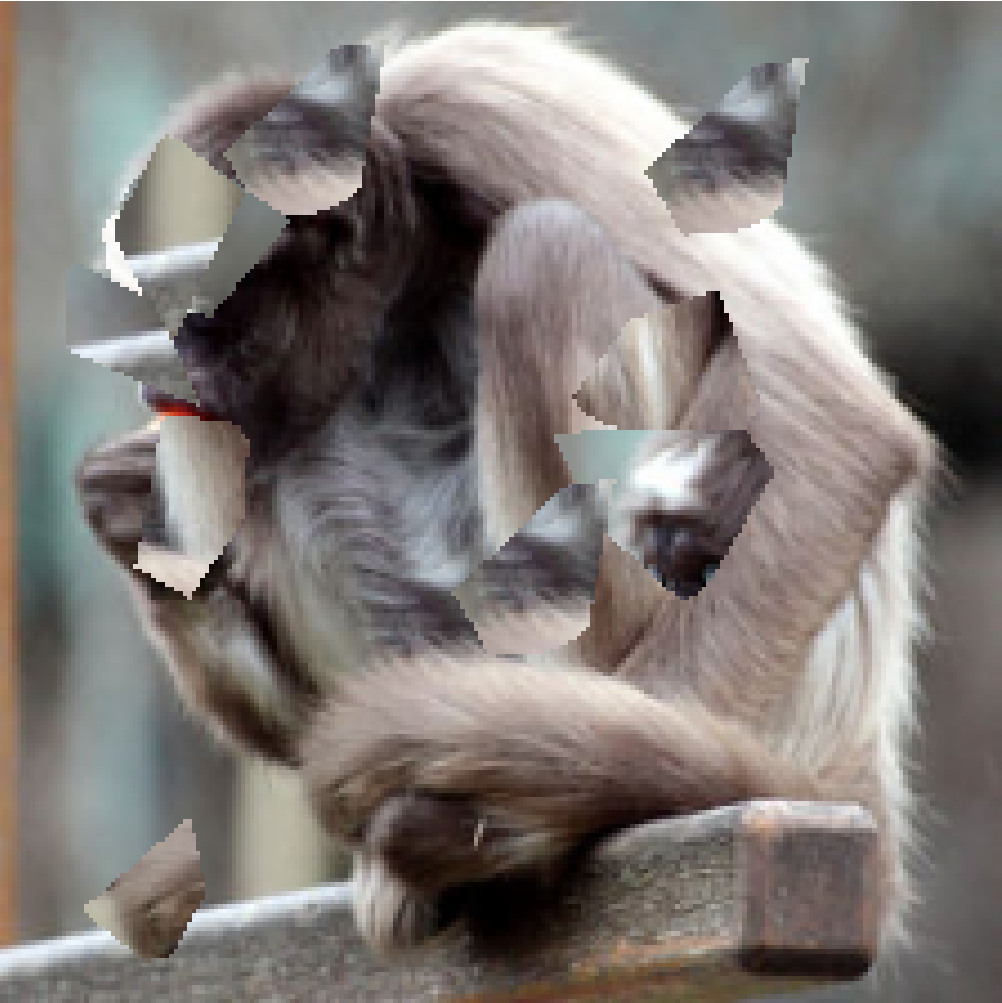}} \hfil
    \subfloat{\includegraphics[width=0.32\linewidth]{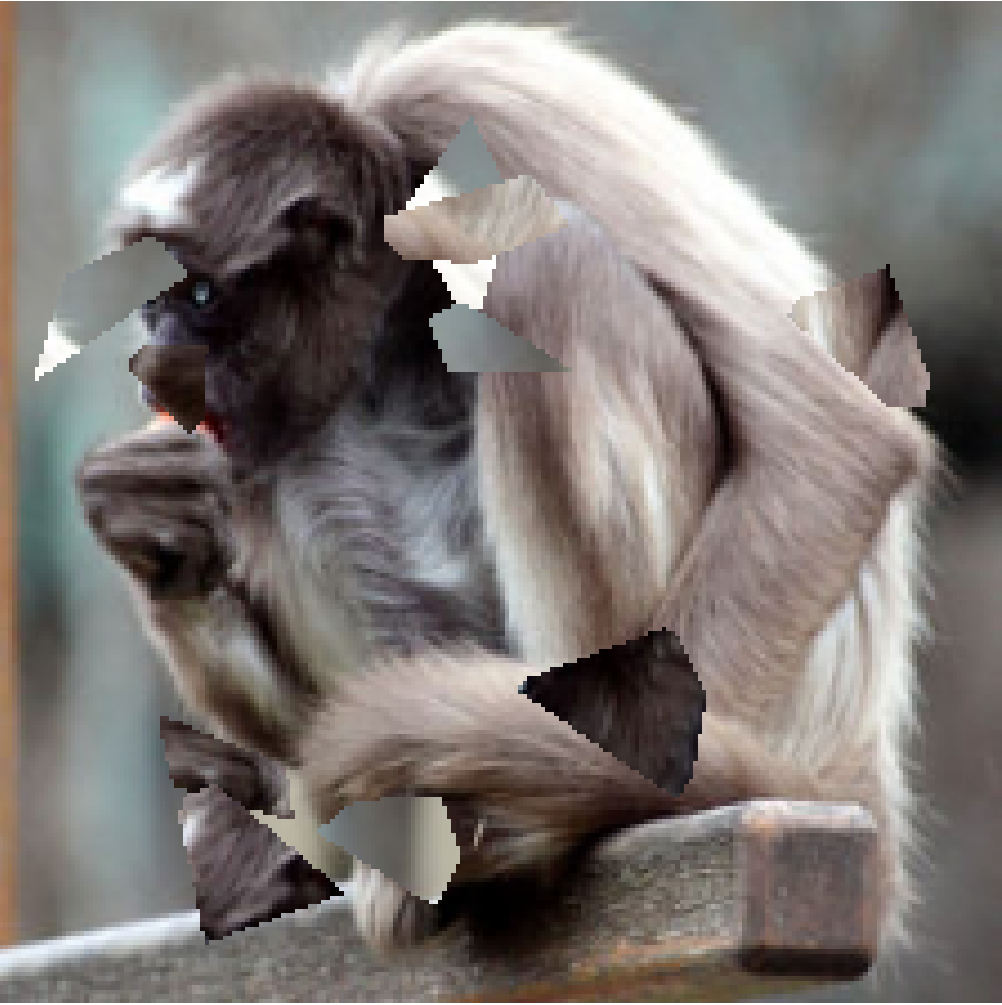}}
    \caption{ Comp. of patch sizes using (left) 50, (middle) 70, and (right) 90 gen. points (at 10 patches, smoothing).}
    \label{fig:patch_examples}
\end{figure}

The variance in patch sizes is also determined by how generator points are distributed across the image. 
By visual inspection, a trend of smaller patches resulting from more generator points, as well as, larger patches resulting from fewer generator points can be discovered. 
Furthermore,~\autoref{fig:patch_examples} shows that depending on the size of an object in an image, a patch may occlude or duplicate a larger or smaller feature. 
Patches generated with 50 generator points in the left-hand column of~\autoref{fig:patch_examples} are large enough to only contain a smaller feature like the face or hand of the monkey, but large enough to contain a large feature from the insect like its head or thorax. 

\section{\uppercase{Conclusion}}
\label{sec:conclusion}
\vspace{-6pt}
In this work we introduced \acf{vp}, a novel data augmentation method and category (transport), to solve the problem of overfitting for CNNs.
We sought to minimize information loss and pixel artifacts, as well as exploit non-orthogonal shapes and structures in data augmentation.
Our method, which employs non-orthogonal shapes and structures to re-combine information within an image, outperforms the existing \ac{da} methods regarding model variance and overfitting tendencies.
Additional experiments analyzed VoronoiPatches' influence on predictions, as well as the influence of \ac{hp} values on performance. 
We show that there are further opportunities to build on our findings and add validity to our initial evaluation of \ac{vp}. 
This includes: optimizing smooth transitions, exploring different pixel values or patch shapes (e.g., black pixels, or small squares or rectangles) to better understand the contribution to data augmentation for CNN, mixing images across the training set, and eventually exploring applications of \ac{vp} in other tasks or fields (e.g., medical image analysis or the field of audio in the form of mel spectrograms). 
From a practical standpoint, solving the limitation of expensive training would enable more efficient usage of the available data sets. 
Currently, expensive training is a limitation of VoronoiPatches, which also is a possible future task.

\newpage
\bibliographystyle{apalike}
{\small
\bibliography{main}}

\end{document}